\documentclass{article}
\usepackage{iclr2020_conference,times}

\usepackage{graphicx}
\usepackage{subfigure}
\definecolor{airforceblue}{rgb}{0.36, 0.54, 0.66}
\usepackage[colorlinks=true, citecolor=airforceblue, linkcolor=airforceblue, urlcolor=airforceblue,filecolor=airforceblue ,menucolor=airforceblue ,runcolor=airforceblue]{hyperref}

\usepackage{amsmath,amsfonts,bm}

\def\eqref#1{equation~\ref{#1}}

\def\1{\bm{1}}

\DeclareMathAlphabet{\mathsfit}{\encodingdefault}{\sfdefault}{m}{sl}
\SetMathAlphabet{\mathsfit}{bold}{\encodingdefault}{\sfdefault}{bx}{n}

\usepackage{hyperref}
\usepackage{url}

\title{CNN-based Approach for Cervical Cancer Classification in Whole-Slide Histopathology Images}

\author{Ferdaous Idlahcen \\
LIMIARF, Faculty of Sciences \\
Mohammed V University \\
Rabat, Morocco \\
\texttt{ferdaous\_idlahcen@um5.ac.ma} \\
\And
Mohammed Majid Himmi \\
LIMIARF, Faculty of Sciences \\
Mohammed V University \\
Rabat, Morocco \\
\texttt{mohammed-majid.himmi@um5.ac.ma} \\
\And
Abdelhak Mahmoudi \\
LIMIARF \\
École Normale Supérieure \\
Mohammed V University \\
Rabat, Morocco \\
\texttt{abdelhak.mahmoudi@um5.ac.ma}
}

\iclrfinalcopy 

\begin{document}

\maketitle

\begin{abstract}

Cervical cancer will cause 460 000 deaths per year by 2040, approximately 90\% are Sub-Saharan African women. A constantly increasing incidence in Africa making cervical cancer a priority by the World Health Organization (WHO) in terms of screening, diagnosis, and treatment. Conventionally, cancer diagnosis relies primarily on histopathological assessment, a deeply error-prone procedure requiring  intelligent computer-aided systems as low-cost patient safety mechanisms but lack of labeled data in digital pathology limits their applicability. In this study, few cervical tissue digital slides from TCGA data portal were pre-processed to overcome whole-slide images obstacles and included in our proposed VGG16-CNN classification approach. Our results achieved an accuracy of 98,26\% and an F1-score of 97,9\%, which confirm the potential of transfer learning on this weakly-supervised task.

\end{abstract}

\section{Introduction}

Cervical cancer (CC) arises from the cervix. The pathogenesis involves human papillomavirus (HPV) in 99,7\% of cases, with 71\% belonging to two high-risk HPVs (HR-HPV) genotypes: HPV-16 and HPV-18~\citep{ngoma_cancer_2019, khazaei_comparison_2016}. Histologically, cervical cancer has two major types, squamous cell carcinoma (SCC) from the squamous epithelium of the ectocervix, and adenocarcinoma (AC) from the glandular epithelium of the endocervix~\citep{ngoma_cancer_2019}. Routinely, only a microscopic examination of biopsy tissue samples by a board-certified pathologist can confirm the cancer diagnosis~\citep{Lee2019}. This pathologist interpretation of tumor histology provides a set of decisive information (i.e. stage, type, choice of treatment, etc.), which is very challenging and error-prone (e.g. human eye-brain system limitations, fatigue and distraction factors, non-informative regions, etc.)~\citep{Lee2019, petrick_evaluation_2013}.  

Currently, tissue specimens have shifted from conventional glass slides to whole-slide images (WSIs), the emergence of this virtual microscopy is increasing the use of deep convolutional neural networks (CNNs) in digital pathology despite WSIs conformation (e.g. computational expenses, single WSI dimensions, lack of annotation, etc.)~\citep{gutman_digital_2017,tizhoosh_artificial_2018}.   

The major contribution of this paper focuses on the potential of a pre-trained CNN approach for hematoxylin \& eosin (H\&E) histopathology image analysis of cervical cancer to acquire an efficient classification despite unlabeled and data poorness problems. Hence, the challenge relied on the non-trivial task of WSIs processing and the VGG16-CNN pre-trained model fine-tuning from a totally different data on which it was trained on.

\section{Materials}

\subsection{Dataset Acquisition}
Through Genomic Data Commons (GDC) data portal, we retrieved histopathology WSIs of SCC and AC H\&E-stained biopsy samples from the Cancer Genome  Atlas (TCGA) database as Aperio SVS files. TCGA database encompasses a huge amount of WSIs for over 30~human tumors at more than 60~primary sites, along with associated omics, radiology, and clinical-outcome data~\citep{tomczak_cancer_2015}. In this study, the 10~experimental histopathology images consist of two common tumor types of the uterine cervix: 5~SCC and 5~AC. 

Each slide used belongs to a single patient.

\subsection{Inclusion Criteria}
\label{Inclusion Criteria}
Formalin-fixed paraffin-embedded (FFPE) tissue specimens are the gold standard for diagnostic medicine, while frozen tissue specimens are suitable for genomic analysis~\citep{gutman_digital_2017}. Since computational pathology requires high-quality input images, we selected only FFPE WSIs due to the good tissue morphology safeguarding. The identification was carried out using relevant metadata embedded in TCGA barcodes~\citep{tomczak_cancer_2015}. 

\begin{figure}[h]
\centering
\subfigure[Tissue sample of patient id-A3HU outlined and ink-stained with a red marker; the white lines across are cutting artifacts.]{\label{fig. 1: a}\includegraphics[width=42mm, height=61mm]{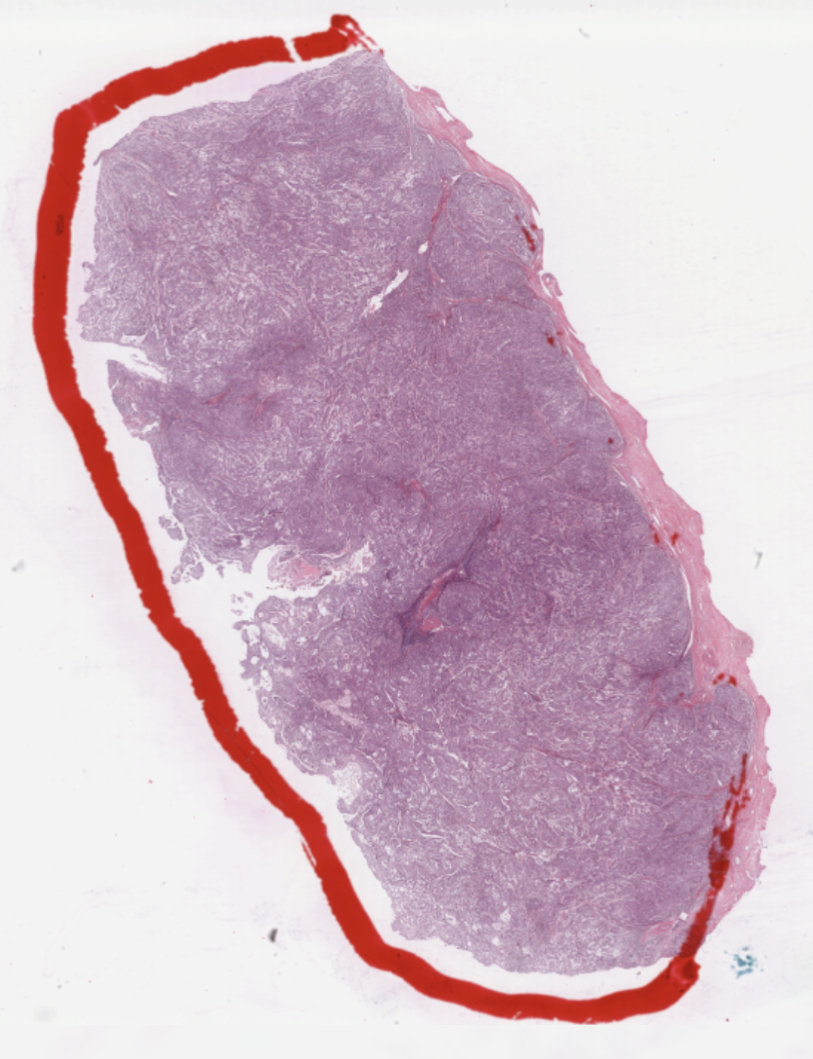}}
\hspace{4mm}
\subfigure[Tissue sample of patient id-A20A having tissue-folds on the top left area, blue ink residues, air bubbles, and cutting artifacts.]{\label{fig. 1: b}\includegraphics[width=42mm, height=61mm]{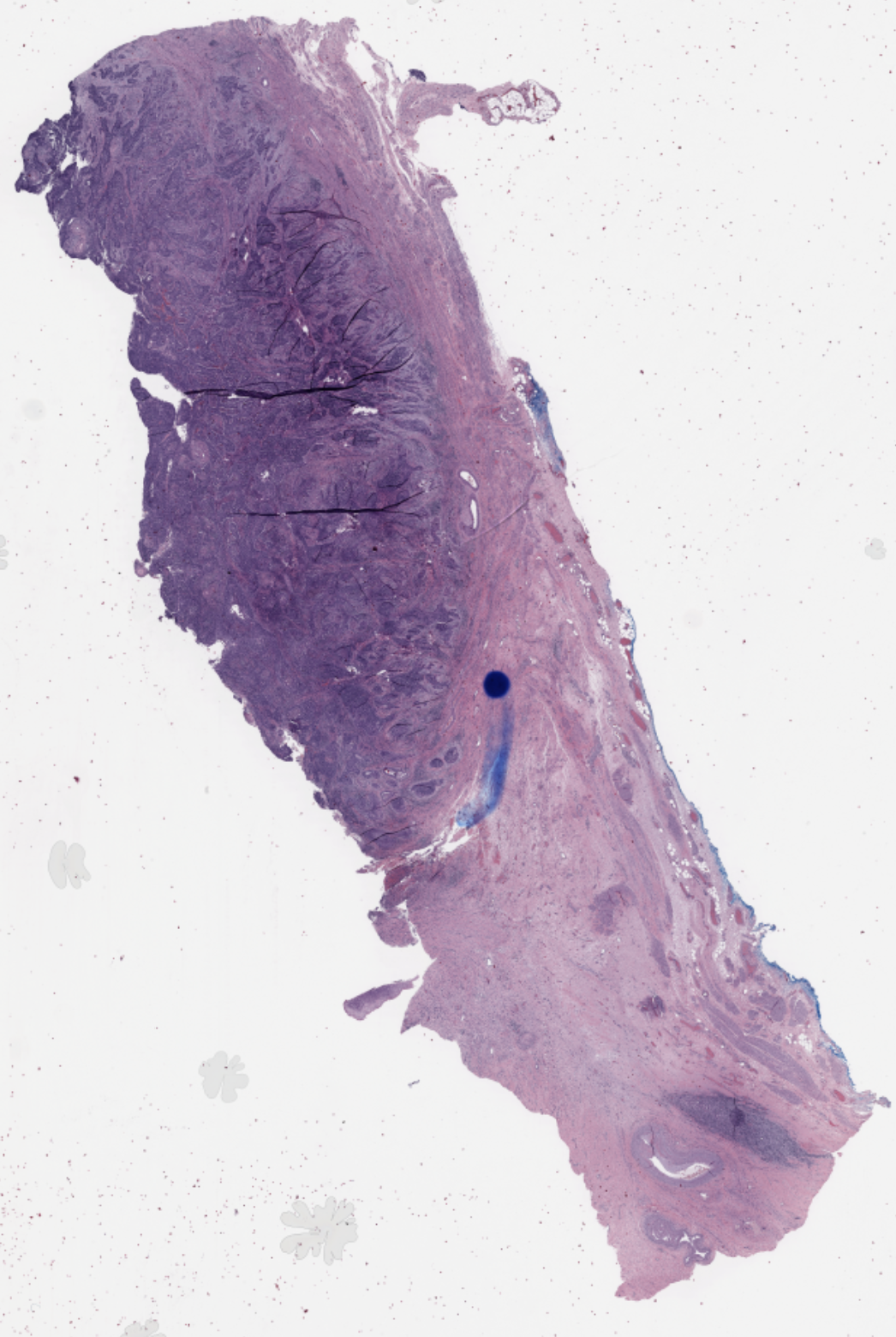}}
\hspace{4mm}
\subfigure[Tissue sample of patient id-A5QV with a significant air bubble and heavily stained in some regions due to tissue-folds.]{\label{fig. 1: c}\includegraphics[width=42mm, height=61mm]{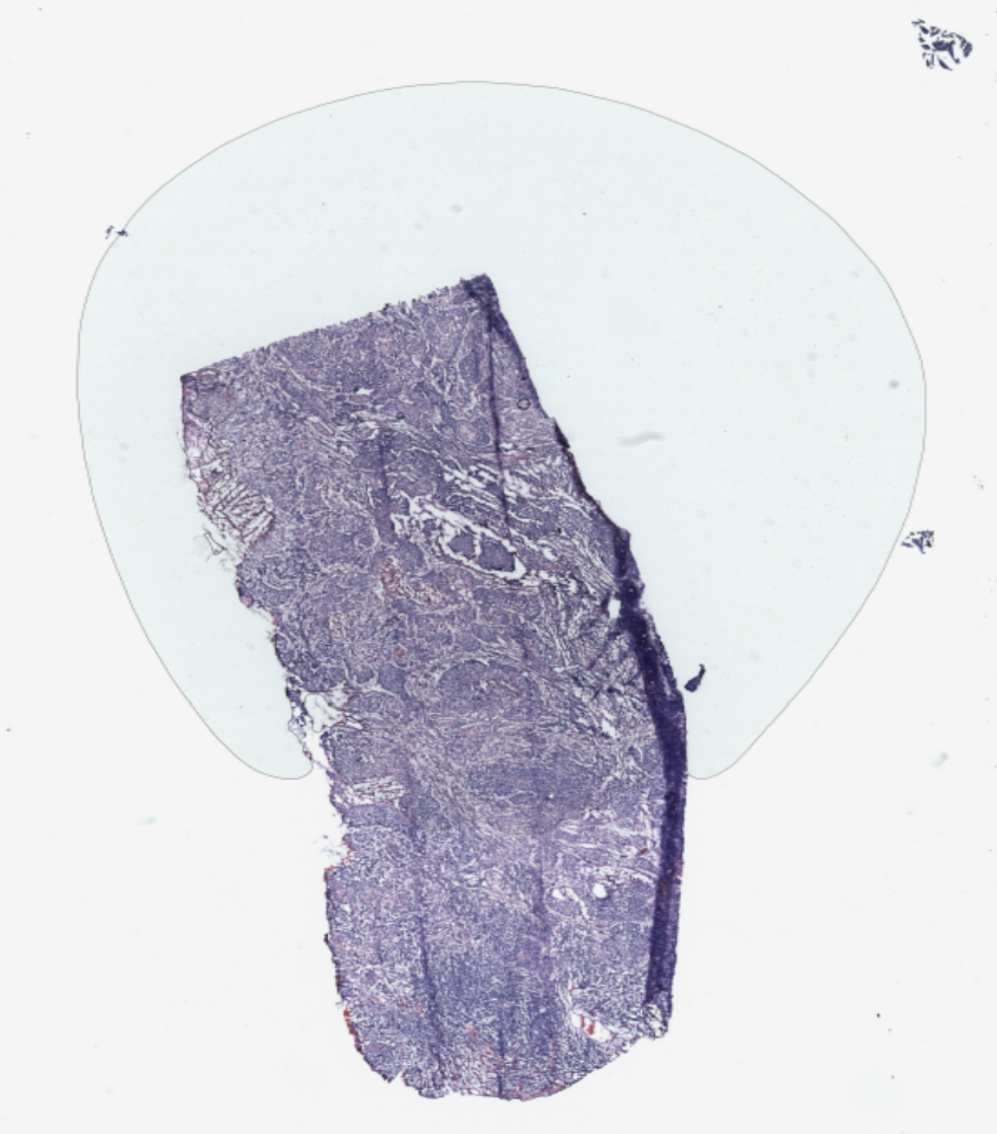}}
\caption{Examples of artifacts in WSIs adopted from GDC data portal.}
\label{fig. 1}
\end{figure}

Moreover, tissue-biopsy slides unveil multiple batch effects and artifacts as a result of alterations related to surgical removal, tissue-processing, microtomy, digitization, etc.~\citep{taqi_review_2018}, some examples are shown in Fig. \ref{fig. 1}. These compromising issues may be misapprehended unintentionally by pathologists and subsequently computer-based learning algorithms~\citep{Dimitriou2019}. Hence, we controlled each slide manually as closely as possible for an improved WSI-based classification. 

We performed additional quality control using \verb+HistoQC+~\citep{janowczyk_histoqc:_2019}.

\subsection{Stain Normalisation}
The histopathological diagnosis is largely based on the H\&E staining protocol. Hematoxylin is a nuclear dye, it colors cell nuclei in blue or deep purple, while eosin is a cytoplasmic dye, it stains cytoplasm and other basic cellular elements in pink or red~\citep{fischer_hematoxylin_2008}. The stained tissues exhibit color variations that may not affect the pathologist's interpretation but have serious implications in subsequent computational medical analysis.

So far, we already discarded low stain condition WSIs during the quality control phase as mentioned in Section~\ref{Inclusion Criteria}. Further, we applied the~\citeauthor{inproceedings} method to better represent the pathological information of the samples.

\subsection{WSI Tile-Based Approach}
The slides are typically scanned at high magnifications ranging from 2.5x to 40x, so an average SVS-format WSI is significantly larger than the other digital formats and requires, therefore, regions of interest (ROIs) extraction~\citep{gutman_cancer_2013}. First, we scaled all the images to 20x magnification with bi-cubic interpolation~\citep{reza_glioma_2016}. Then, we tiled each WSI file into non-overlapping patches.

Our tiling method was centered while ignoring the remainder edge tiles to create 1024x1024~pixels images at 20x, which is the same size/magnification used by a trained pathologist for a whole tumor report~\citep{article}. Eventually, we extracted 12 412~tiles of 1024x1024~pixels from the 10~WSIs.

Another key step is ascribable to the irrelevant white space in non-tissue regions where non-proliferating structures are visible. This process is often difficult because extracting ROIs is usually manually-operated. Accordingly, we removed only full white tiles automatically using red green blue (RGB) values technique but the rest was done customarily.

In the end, 300~SCC and 300~AC tiles containing at least 90\% of tissue and one distinguishing histological criterion (e.g. keratin pearl, atypical mitosis, etc.) were chosen for further training.

\section{Methods and Results}

In our experiment, we used VGG-16 architecture as presented in Table~\ref{table. 1}. All images were resized to 224-by-224 and split as follows: 70\% for train, 10\% for validation, and 20\% for evaluation. Table~\ref{table. 2} summarizes the number of SCC and AC tiles in each set.  

\begin{table}[h!]
\caption{Architecture of VGG-16 network.}
\vspace{1em}
\label{table. 1}
\begin{center}
\begin{tabular}{c|c}
\textbf{Layer}   &\textbf{Size}
\\ \hline \\
Conv~x2		&224 x 224 x 64  \\
Pool		&112 x 112 x 64 \\
Conv~x2		&112 x 112 x 128\\
Pool		&56 x 56 x 128\\
Conv~x3		&56 x 56 x 256\\
Pool		&28 x 28 x 256\\
Conv~x3		&28 x 28 x 512\\
Pool		&28 x 28 x 512 \\
Conv~x3		&14 x 14 x 512\\
Pool		&7 x 7 x 512\\
FC		&25088\\
FC		&4096\\
FC		&4096\\
\end{tabular}
\end{center}
\end{table}

\begin{table}[h!]
\caption{Data split statistics.}
\label{table. 2}
\vspace{1em}
\centering
\begin{tabular}{c|c|c}
\textbf{Train} & \textbf{Validation} & \textbf{Test}\\ \hline
216~SCC &24~SCC  &60~SCC\\
216~AC &24~AC &60~AC
\end{tabular}
\end{table}

Firstly, we extracted and stored features from the model's last convolution layer (7, 7, 512) with ImageNet pre-trained weights. Secondly, the deep learning extracted features were used effectively in a subsequent fine-tuned VGG-16 classifier after achieving a 97\% accuracy. 

In the fine-tuning process, we froze the first four conv blocks and redefined the last convolutional one as well as the fully-connected layers. As main strategies, we used stochastic gradient descent (SGD) as optimizer and binary cross-entropy as loss function for 30~epochs in total. The training was performed with an initial learning rate value of 0.0001 and a batch size of 16.

As shown in Table~\ref{table. 3}, we evaluated the classification using four metrics: accuracy, precision, recall, and F1 score. Our final submission task achieved an accuracy of 98,26\% and an F1 score of 97,9\% for the binary SCC versus AC output, which is perfectly sufficient for practical use. We assume that these results can be further improved by integrating informative labeled ROIs features or deep generative models.

\begin{table}[h!]
\caption{Performance on the test set.}
\label{table. 3}
\vspace{1em}
\centering
\begin{tabular}{c|c|c|c}
\textbf{Accuracy (\%)} & \textbf{Precision (\%)} & \textbf{Recall (\%)} & \textbf{F1 (\%)}\\ \hline
98.26 &96.8  &99 &97.9
\end{tabular}
\end{table}

\section{Conclusion}

For the first time, a pre-trained CNN-VGG16 approach was implemented for H\&E-stained histopathological uterine cervix tumors classification. The dataset was provided from TCGA data portal and underwent multiple pre-process methods to overcome WSIs conformation obstacles. Our experimental findings are remarkable and demonstrate the capability of deep learning pre-trained models toward promising computer-aided diagnosis in digital pathology. Eventually, this study is an initiation to other researches in light of the fact that some cancer patterns can not be gleaned by human examination. In addition, there is a clear potential of transfer learning in histo-genomics correlations on which we intend to work.

\bibliography{iclr2020_conference}
\bibliographystyle{iclr2020_conference}

\end{document}